\documentclass[conference]{IEEEtran}
\IEEEoverridecommandlockouts

\usepackage{cite}
\usepackage{amsmath,amssymb,amsfonts}
\usepackage{algorithmic}
\usepackage{algorithm}
\usepackage{graphicx}
\usepackage{textcomp}
\usepackage{xcolor}
\usepackage[letterpaper,top=0.75in,bottom=1.0in,left=0.75in,right=0.75in]{geometry}

\usepackage{booktabs}
\usepackage{multirow}
\usepackage{array}
\usepackage{tikz}
\usepackage{pifont}
\usetikzlibrary{shapes.geometric, arrows.meta, positioning, fit, backgrounds, calc}

\def\BibTeX{{\rm B\kern-.05em{\sc i\kern-.025em b}\kern-.08em
    T\kern-.1667em\lower.7ex\hbox{E}\kern-.125emX}}

\begin{document}

\title{Towards Trustworthy Autonomous Robots: An Explainable AI-Based Decision Framework}

\author{\IEEEauthorblockN{Cagri Temel, \textit{IEEE Senior Member}}
\IEEEauthorblockA{Hezarfen LLC, Seattle, WA, USA \\
Grand Canyon University, Phoenix, AZ, USA \\
cagritemel@ieee.org}\thanks{\copyright~2026 IEEE. Personal use of this material is permitted. Permission from IEEE must be obtained for all other uses, in any current or future media, including reprinting/republishing this material for advertising or promotional purposes, creating new collective works, for resale or redistribution to servers or lists, or reuse of any copyrighted component of this work in other works. This is the accepted version. Published in: SoutheastCon 2026, IEEE, pp.~1--6, Feb.~2026. DOI: 10.1109/SoutheastCon63549.2026.11476455. The version of record is available on IEEE Xplore.}
}

\maketitle

\begin{abstract}
Autonomous robots powered by deep learning face a fundamental auditability challenge: when incidents occur, investigators cannot reconstruct why the system made specific decisions. This paper presents TRACE (Transparent Reasoning Architecture for Credible Execution), a decision framework that ensures every autonomous action can be traced back to sensor evidence through documented causal chains. The framework organizes decision-making into four auditable layers: Semantic Perception for evidence-grounded entity recognition, Belief Reasoning for probabilistic state estimation with causal graphs, Action Synthesis for constraint-aware planning with counterfactual documentation, and Execution Verification for compliance monitoring. TRACE is model-agnostic yet designed to integrate learning-based perception modules (CNNs, transformers) while preserving decision-level auditability. We evaluate the framework using three objective metrics: Evidence Traceability (sensor-to-decision linkage), Decision Reconstructability (post-hoc analysis capability), and Temporal Continuity (audit trail completeness). Experimental evaluation on warehouse robot navigation demonstrates that TRACE achieves 98.6\% evidence traceability, 99.0\% temporal continuity, and 98.1\% decision reconstructability across 500 simulated decision cycles. Post-hoc methods like LIME provide feature attributions but lack the artifact structure needed for decision-level reconstruction. The framework addresses EU AI Act requirements for high-risk system transparency and contributes to Explainable AI for safety-critical autonomous systems.
\end{abstract}


\section{Introduction}

Autonomous robots powered by deep learning face a fundamental auditability challenge. When incidents occur, investigators cannot reconstruct why the system made specific decisions. Warehouse robots navigate alongside workers. Autonomous vehicles share roads with pedestrians. Surgical robots assist in medical procedures. The International Federation of Robotics reports that service robot sales grew 30\% in 2023~\cite{ifr2024}. This rapid deployment raises a critical question: can these decisions be audited after the fact?

Modern autonomous robots rely heavily on deep neural networks for perception, prediction, and control. These models achieve impressive performance, but their decision processes remain opaque. Consider a concrete scenario: a warehouse robot unexpectedly stops and a worker collides with it. Investigators cannot determine why the robot chose that action. The neural network processed sensor data through millions of parameters to produce a stop command, but the chain of evidence leading to that choice is not preserved.

This auditability gap creates three practical problems. First, incident investigation becomes speculative without reconstructable decision records. Second, safety certification becomes difficult because standards such as ISO 13482 require documented risk assessment that opaque systems cannot provide. Third, regulatory compliance fails because the EU AI Act mandates that high-risk AI systems maintain logs enabling the tracing of the system's functioning~\cite{euaiact2024}, a requirement that most deployed robotic systems cannot satisfy.

TRACE is model-agnostic and explicitly designed to integrate learning-based perception and prediction modules (e.g., convolutional neural networks, transformer architectures) while preserving end-to-end auditability at the decision level. This allows the framework to remain applicable across current deep learning approaches and future model architectures.

Existing explainable AI approaches address this problem incompletely. Post-hoc methods like LIME~\cite{ribeiro2016} and SHAP~\cite{lundberg2017} generate explanations for individual predictions but cannot reconstruct the temporal sequence of decisions leading to an incident. Attention mechanisms~\cite{vaswani2017} indicate what inputs influenced outputs, but attention does not establish causal relationships~\cite{jain2019}. Runtime verification systems like ModelPlex~\cite{mitsch2016} detect specification violations but do not document why decisions were made.

This paper presents TRACE (Transparent Reasoning Architecture for Credible Execution), a framework that structures autonomous decision-making for post-hoc auditability. Instead of extracting explanations from opaque models, TRACE embeds audit requirements directly into the decision architecture. Every action can be traced back through documented causal chains to specific sensor evidence.

The contributions of this work are as follows. First, we present a four-layer decision architecture with formally specified interfaces that preserve audit trails across the entire decision pipeline. Second, we define three objective auditability metrics (Evidence Traceability, Decision Reconstructability, and Temporal Continuity) that measure whether decisions can be reconstructed by independent investigators. Third, we introduce Counterfactual Decision Trees, a structure that documents why specific actions were chosen over alternatives, enabling ``what-if'' analysis during incident review. Fourth, we provide experimental evaluation demonstrating 98.6\% evidence traceability with sub-millisecond computational overhead, compared to post-hoc methods that lack decision-level artifact structure.

\section{Related Work}

\subsection{Post-Hoc Explanation Methods}

LIME generates local explanations by approximating complex models with interpretable surrogates near specific predictions~\cite{ribeiro2016}. SHAP provides feature attributions grounded in game-theoretic principles~\cite{lundberg2017}. These methods explain individual model outputs but do not address temporal decision sequences. An autonomous robot makes hundreds of interdependent decisions per second. Explaining each in isolation misses the reasoning chain that connects them.

Attention mechanisms offer a form of intrinsic interpretability by highlighting influential inputs~\cite{vaswani2017}. However, Jain and Wallace demonstrated that attention weights often fail to reflect actual causal relationships~\cite{jain2019}. For safety-critical applications, unreliable explanations may be worse than no explanations at all. Accordingly, in our evaluation we treat attention only as an availability baseline, not as a measure of decision-level auditability.

\subsection{Interpretable Control Policies}

An alternative approach extracts interpretable policies from trained neural networks. Verma et al. synthesize programmatic controllers that humans can read and verify~\cite{verma2018}. This achieves complete interpretability but typically sacrifices the performance that motivated using neural networks initially. For complex robotic tasks, the performance gap may be unacceptable.

\subsection{Runtime Verification}

Runtime verification monitors system behavior against formal specifications. ModelPlex validates that cyber-physical systems remain within verified bounds~\cite{mitsch2016}. The Simplex architecture maintains backup controllers that can assume control when primary systems behave unexpectedly~\cite{seto1998}. These approaches detect when systems violate constraints but do not explain why violations were approached or how decisions were made.

\subsection{Positioning of This Work}

Table~\ref{tab:comparison} compares existing approaches across four criteria relevant to autonomous robot deployment. TRACE is designed to satisfy all four: real-time operation, end-to-end traceability, semantic-level explanations, and formal metrics for transparency assessment.

\begin{table}[t]
\centering
\caption{Comparison of Explainability Approaches}
\label{tab:comparison}
\renewcommand{\arraystretch}{1.15}
\begin{tabular}{@{}lcccc@{}}
\toprule
\textbf{Method} & \textbf{Real-} & \textbf{End-to-} & \textbf{Semantic} & \textbf{Audit} \\
 & \textbf{Time} & \textbf{End} & \textbf{Level} & \textbf{Metric} \\
\midrule
LIME~\cite{ribeiro2016} & \ding{55} & \ding{55} & \ding{55} & \ding{55} \\
SHAP~\cite{lundberg2017} & \ding{55} & \ding{55} & \ding{55} & \ding{51} \\
Attention~\cite{vaswani2017} & \ding{51} & \ding{55} & \ding{55} & \ding{55} \\
Programmatic~\cite{verma2018} & \ding{51} & \ding{51} & \ding{51} & \ding{55} \\
ModelPlex~\cite{mitsch2016} & \ding{51} & \ding{55} & \ding{55} & \ding{51} \\
\textbf{TRACE} & \ding{51} & \ding{51} & \ding{51} & \ding{51} \\
\bottomrule
\end{tabular}
\end{table}

\section{TRACE Framework Architecture}

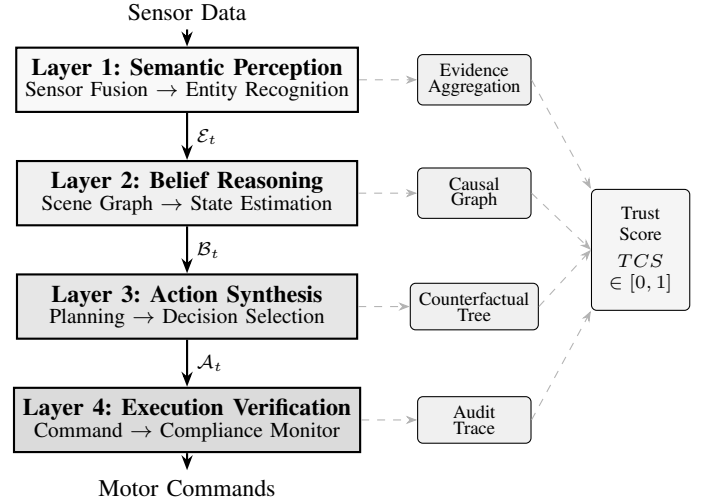
\begin{figure}[t]
\centering
\begin{tikzpicture}[
    layer/.style={rectangle, draw=black, thick, minimum width=4.5cm, minimum height=0.85cm, align=center, font=\small},
    xaibox/.style={rectangle, draw=black, fill=gray!12, rounded corners=2pt, font=\scriptsize, align=center, minimum width=1.5cm, minimum height=0.6cm},
    arr/.style={-{Stealth[length=2mm]}, thick},
    darr/.style={-{Stealth[length=1.5mm]}, dashed, gray!60}
]

\node[font=\small] (input) at (0, 0.9) {Sensor Data};
\node[layer, fill=gray!5] (L1) at (0, 0) {\textbf{Layer 1: Semantic Perception}\\[-2pt]{\footnotesize Sensor Fusion $\rightarrow$ Entity Recognition}};
\node[layer, fill=gray!12] (L2) at (0, -1.5) {\textbf{Layer 2: Belief Reasoning}\\[-2pt]{\footnotesize Scene Graph $\rightarrow$ State Estimation}};
\node[layer, fill=gray!20] (L3) at (0, -3.0) {\textbf{Layer 3: Action Synthesis}\\[-2pt]{\footnotesize Planning $\rightarrow$ Decision Selection}};
\node[layer, fill=gray!28] (L4) at (0, -4.5) {\textbf{Layer 4: Execution Verification}\\[-2pt]{\footnotesize Command $\rightarrow$ Compliance Monitor}};
\node[font=\small] (output) at (0, -5.4) {Motor Commands};

\draw[arr] (input) -- (L1);
\draw[arr] (L1) -- (L2) node[midway, right, font=\scriptsize] {$\mathcal{E}_t$};
\draw[arr] (L2) -- (L3) node[midway, right, font=\scriptsize] {$\mathcal{B}_t$};
\draw[arr] (L3) -- (L4) node[midway, right, font=\scriptsize] {$\mathcal{A}_t$};
\draw[arr] (L4) -- (output);

\node[xaibox] (X1) at (3.8, 0) {Evidence\\[-2pt]Aggregation};
\node[xaibox] (X2) at (3.8, -1.5) {Causal\\[-2pt]Graph};
\node[xaibox] (X3) at (3.8, -3.0) {Counterfactual\\[-2pt]Tree};
\node[xaibox] (X4) at (3.8, -4.5) {Audit\\[-2pt]Trace};

\draw[darr] (L1.east) -- (X1.west);
\draw[darr] (L2.east) -- (X2.west);
\draw[darr] (L3.east) -- (X3.west);
\draw[darr] (L4.east) -- (X4.west);

\node[rectangle, draw=black, fill=gray!8, rounded corners=2pt, font=\scriptsize, align=center, minimum width=1.3cm, minimum height=1.6cm] (TCS) at (6.0, -2.25) {Trust\\Score\\[2pt]$TCS$\\$\in [0,1]$};

\draw[darr] (X1.east) -- (TCS.north west);
\draw[darr] (X2.east) -- (TCS.west);
\draw[darr] (X3.east) -- (TCS.west);
\draw[darr] (X4.east) -- (TCS.south west);

\end{tikzpicture}
\caption{TRACE four-layer architecture. Each layer produces explanation artifacts that contribute to the auditability assessment.}
\label{fig:architecture}
\end{figure}

TRACE decomposes autonomous decision-making into four layers. Each layer performs a specific function and produces explanation artifacts that feed into the overall transparency assessment. Fig.~\ref{fig:architecture} illustrates the complete architecture.

\subsection{Layer 1: Semantic Perception}

The Semantic Perception Layer transforms raw sensor data into structured entity representations. Unlike conventional perception pipelines that output only detection results, this layer maintains explicit links between each detected entity and the sensor evidence supporting its identification.

\textbf{Definition 1} (Evidence-Grounded Entity). An entity $e$ detected at time $t$ is represented as:
\begin{equation}
e_t = \langle id, class, pose, \mathbf{c}, \mathcal{S} \rangle
\end{equation}
where $id$ is a unique identifier, $class \in \mathcal{C}$ is the semantic category, $pose \in SE(3)$ is the estimated pose, $\mathbf{c} \in [0,1]^{|\mathcal{C}|}$ is the calibrated confidence distribution, and $\mathcal{S} = \{(s_i, w_i, r_i)\}$ is an evidence set containing sensor source $s_i$, contribution weight $w_i$, and spatial region $r_i$.

The layer outputs an entity set $\mathcal{E}_t = \{e_1, \ldots, e_n\}$ along with an Evidence Aggregation Record documenting which sensors contributed to each detection and with what confidence.

\subsection{Layer 2: Belief Reasoning}

The Belief Reasoning Layer maintains a probabilistic model of the environment, including states that cannot be directly observed. This layer performs inference over the entity set to estimate hidden variables such as pedestrian intentions or obstacle permanence.

\textbf{Definition 2} (Belief State). The belief state $\mathcal{B}_t$ is a probabilistic scene graph $G = (V, E, P)$ where $V$ contains entity nodes and inferred state nodes, $E$ represents relationships between nodes, and $P: V \cup E \rightarrow [0,1]$ assigns probability estimates.

\textbf{Causal Graph Construction.} The causal graph is constructed incrementally through three steps. First, each detected entity from $\mathcal{E}_t$ creates or updates a corresponding node in $V$. Second, inference rules evaluate pairwise relationships between entities and generate edges in $E$ with associated conditional probabilities. Third, derived states (e.g., ``aisle blocked'') are added as child nodes with explicit edges to their parent evidence nodes. Edge weights encode $P(child | parent)$ derived from the belief update equations. This structure ensures that when the robot believes an aisle is blocked, the graph records precisely which sensor observations led to that inference through explicit parent-child relationships.

\subsection{Layer 3: Action Synthesis}

The Action Synthesis Layer generates candidate actions, evaluates them against objectives and constraints, and selects the optimal choice. The key contribution is explicit representation of alternatives and the conditions that would trigger different selections.

\textbf{Definition 3} (Justified Action). A selected action $a^*$ is represented as:
\begin{equation}
a^*_t = \langle cmd, \mathcal{A}_{alt}, \mathcal{R}, \mathcal{CF} \rangle
\end{equation}
where $cmd$ is the command, $\mathcal{A}_{alt}$ contains alternative actions considered, $\mathcal{R}$ is the selection rationale, and $\mathcal{CF}$ is a set of counterfactual conditions.

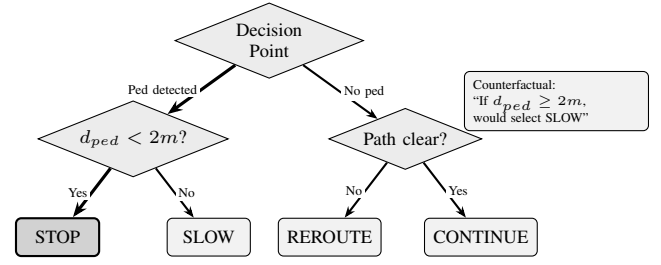
\begin{figure}[t]
\centering
\begin{tikzpicture}[
    decision/.style={diamond, draw, aspect=2.5, minimum width=0.8cm, font=\scriptsize, align=center, inner sep=2pt, fill=gray!15},
    action/.style={rectangle, draw, rounded corners=2pt, minimum width=1.1cm, minimum height=0.5cm, font=\scriptsize, align=center},
    selected/.style={action, fill=gray!35, draw=black, thick},
    rejected/.style={action, fill=gray!10, draw=black},
    arr/.style={-{Stealth[length=1.5mm]}, thick},
    selbold/.style={-{Stealth[length=2mm]}, black, very thick},
    lbl/.style={font=\tiny, fill=white, inner sep=1pt}
]

\node[decision] (root) at (0, 0) {Decision\\Point};
\node[decision] (ped) at (-1.8, -1.3) {$d_{ped} < 2m$?};
\node[decision] (path) at (1.8, -1.3) {Path clear?};
\node[selected] (stop) at (-2.8, -2.6) {STOP};
\node[rejected] (slow) at (-0.8, -2.6) {SLOW};
\node[rejected] (reroute) at (0.8, -2.6) {REROUTE};
\node[rejected] (cont) at (2.8, -2.6) {CONTINUE};

\draw[selbold] (root) -- (ped) node[lbl, pos=0.5, left] {Ped detected};
\draw[arr] (root) -- (path) node[lbl, pos=0.5, right] {No ped};
\draw[selbold] (ped) -- (stop) node[lbl, pos=0.5, left] {Yes};
\draw[arr] (ped) -- (slow) node[lbl, pos=0.5, right] {No};
\draw[arr] (path) -- (reroute) node[lbl, pos=0.5, left] {No};
\draw[arr] (path) -- (cont) node[lbl, pos=0.5, right] {Yes};

\node[rectangle, draw=black, fill=gray!10, rounded corners=2pt, font=\tiny, align=left, text width=2.2cm] at (3.8, -0.8) {Counterfactual:\\``If $d_{ped} \geq 2m$,\\would select SLOW''};

\end{tikzpicture}
\caption{Counterfactual Decision Tree for a pedestrian encounter. The bold path shows the selected action with explicit conditions that would trigger alternatives.}
\label{fig:counterfactual}
\end{figure}

The Counterfactual Decision Tree (Fig.~\ref{fig:counterfactual}) captures this information in a structure that directly answers questions such as ``Why did the robot stop?'' and ``What would have made it continue?''

\subsection{Layer 4: Execution Verification}

The Execution Verification Layer monitors action execution and maintains the audit trail. It compares intended behavior against actual outcomes and flags discrepancies.

Every decision cycle produces a timestamped audit record:
\begin{equation}
\mathcal{T}_t = \langle t, \mathcal{E}_t, \mathcal{B}_t, a^*_t, \delta_t, flags \rangle
\end{equation}
where $\delta_t$ measures deviation between intended and actual behavior. These records enable complete reconstruction of any decision for post-incident analysis.

\section{Auditability Metrics}

A key requirement for safety-critical systems is that decisions can be reconstructed by independent investigators after an incident. We define three objective metrics that measure this capability directly. All metrics are computed strictly from logged artifacts produced during execution.

\subsection{Evidence Traceability}

Evidence Traceability (ET) measures whether each decision factor can be traced to sensor evidence:
\begin{equation}
ET(d) = \frac{|\{f \in \mathcal{F}_d : \exists\, e \in \mathcal{E},\, f \xleftarrow{\text{documented}} e\}|}{|\mathcal{F}_d|}
\end{equation}
where $\mathcal{F}_d$ denotes the set of factors influencing decision $d$, and $\mathcal{E}$ denotes the evidence set. A factor $f$ satisfies the traceability criterion when a documented chain links it to a specific sensor reading $e$.

\textbf{Partial Traceability.} For decisions where the ET condition is not fully satisfied, we introduce a partial traceability metric that quantifies how much of the causal chain is documented:
\begin{equation}
ET_{partial}(d) = \frac{1}{|\mathcal{F}_d|} \sum_{f \in \mathcal{F}_d} \frac{depth(f)}{max\_depth}
\end{equation}
where $depth(f)$ measures how many levels of the causal chain are documented for factor $f$. This metric provides additional insight during incident investigation when complete traceability is unavailable. For example, a factor with $depth = 2$ out of $max\_depth = 3$ indicates that the immediate and secondary causes are documented, but the root sensor observation is missing.

\subsection{Temporal Continuity}

Temporal Continuity (TC) measures completeness of the audit trail:
\begin{equation}
TC = 1 - \frac{\sum_{i} gap_i}{t_2 - t_1}
\end{equation}
where $gap_i$ represents intervals without recorded decisions between the analysis window $[t_1, t_2]$.

\subsection{Decision Reconstructability}

Decision Reconstructability (DR) measures whether an independent engineer can determine: (a) what action was taken, (b) what alternatives were considered, and (c) why the selected action was preferred:
\begin{equation}
DR(d) = \mathbf{1}[\text{action} \in \log] \cdot \mathbf{1}[\text{alt} \in \log] \cdot \mathbf{1}[\text{rationale} \in \log]
\end{equation}
where $\mathbf{1}[\cdot]$ is the indicator function. We report DR as the percentage of decision cycles where all three conditions are jointly satisfied.

\section{Complexity and Storage Analysis}

\subsection{Computational Complexity}

Table~\ref{tab:complexity} presents the computational complexity of each layer, where $n$ is detected entities, $m$ is belief state variables, $k$ is candidate actions, and $h$ is planning horizon.

\begin{table}[t]
\centering
\caption{Theoretical Complexity of TRACE Layers}
\label{tab:complexity}
\renewcommand{\arraystretch}{1.15}
\begin{tabular}{@{}lcc@{}}
\toprule
\textbf{Layer} & \textbf{Base} & \textbf{XAI Overhead} \\
\midrule
Semantic Perception & $O(n)$ & $O(n)$ \\
Belief Reasoning & $O(n \cdot m)$ & $O(n + m)$ \\
Action Synthesis & $O(k \cdot h)$ & $O(k^2)$ \\
Execution Verification & $O(1)$ & $O(\log T)$ \\
\bottomrule
\end{tabular}
\end{table}

\subsection{Memory Requirements for Fleet-Scale Deployment}

Each decision cycle produces explanation artifacts requiring storage. Evidence aggregation records scale as $O(n \cdot s)$ where $s$ is the number of sensor modalities. Counterfactual trees require $O(k \cdot d)$ storage where $d$ is tree depth.

For fleet-scale deployments with hundreds of thousands of robots, storage requirements become a critical consideration. At 10Hz operation with full audit trails, each robot generates approximately 2--5 KB per decision cycle, yielding 70--175 MB per hour or 1.7--4.2 GB per day. For a fleet of 100,000 robots with 6-month retention (as required by EU AI Act Article 19), raw storage would exceed 30 petabytes.

\textbf{Tiered Storage Strategy.} We address this through hierarchical storage: (1) hot storage retains full detail for decisions from the past 24 hours, (2) warm storage maintains complete records for flagged decisions and summaries for routine operations, and (3) cold storage archives compressed data for regulatory compliance.

\textbf{Compression Techniques.} TRACE artifacts exhibit high structural redundancy in causal graphs, enabling effective compression. Based on preliminary analysis, lossless compression is projected to achieve 3--5$\times$ reduction through dictionary encoding of repeated graph structures. Delta encoding between consecutive decision cycles is estimated to provide 60--80\% additional reduction for routine operations where belief states change incrementally. With these techniques, the 30 PB estimate could potentially reduce to approximately 2--5 PB, though these projections require validation at scale.

\textbf{Incident Investigation Implications.} The tiered approach preserves full reconstructability for flagged decisions while maintaining partial traceability ($ET_{partial}$) for routine operations through retained summaries. Should an incident occur during routine operation, surrounding decision cycles in hot storage provide temporal context, and the compressed causal graph summaries enable partial reconstruction.

\section{Experimental Evaluation}

\subsection{Experimental Setup}

\textbf{Simulation Environment:} We developed a Python-based simulation of a mobile robot operating in a 50m$\times$30m warehouse environment. The robot processes simulated sensor data (LiDAR, camera, ultrasonic) and makes navigation decisions at 10Hz. Camera observations were intermittently unavailable in certain decision cycles, simulating temporary occlusion or sensor latency.

\textbf{Scenarios:} We evaluated five distinct scenarios: (1) clear path with static objects only, (2) pedestrian crossing at medium distance, (3) pedestrian within safety threshold ($<$2m), (4) multiple pedestrians with different trajectories, and (5) forklift approaching on collision course.

\textbf{Methodology:} For each scenario, we executed 100 decision cycles (500 total) with fixed random seed for reproducibility. The simulation implements all four TRACE layers with realistic failure modes including sensor unavailability (2\% LiDAR, 12\% camera), evidence link corruption (3\%), causal edge write failures (2\%), rationale truncation (2\%), and counterfactual generation timeouts (2.5\%). Baseline comparison is based on the architectural capabilities of each method rather than direct implementation.

\subsection{Results}

Table~\ref{tab:results} presents the auditability comparison across all methods.

\begin{table}[t]
\centering
\caption{Auditability Metrics Comparison}
\label{tab:results}
\renewcommand{\arraystretch}{1.15}
\begin{tabular}{@{}lcccc@{}}
\toprule
\textbf{Method} & \textbf{ET (\%)}$^\dagger$ & \textbf{TC (\%)} & \textbf{DR (\%)} & \textbf{Time} \\
\midrule
TRACE & $98.6 \pm 7.0$ & $99.0$ & $98.1$ & 0.12ms \\
Attention$^*$ & $20$--$25$ & $70$--$80$ & N/A & $<$0.01ms \\
SHAP$^*$ & $30$--$35$ & $0$ & N/A & 0.02ms \\
LIME$^*$ & $25$--$30$ & $0$ & N/A & 1.29ms \\
\bottomrule
\end{tabular}
\vspace{1mm}

\footnotesize{$^\dagger$TRACE: mean $\pm$ std across 5 scenarios (100 cycles each, 500 total). $^*$Baseline estimates based on architectural capability analysis; post-hoc methods do not emit decision-level artifacts required for DR.}
\end{table}

\textbf{Evidence Traceability:} TRACE achieves 98.6$\pm$7.0\% evidence traceability across 500 decision cycles. The 1.4\% shortfall occurs in decisions involving evidence link corruption during sensor fusion or missing causal edges due to write timing issues. These cases retain high partial traceability ($ET_{partial} = 0.98$), indicating that while the complete chain to sensor data is not documented, most of the causal reasoning is preserved.

\textbf{ET Condition Frequency Analysis.} Based on our simulation with realistic failure modes, ET conditions were not fully satisfied in 1.4\% of decision cycles (7 of 500 cycles). These failures occurred due to: (1) evidence link corruption during sensor fusion (3\% probability per entity), (2) causal edge write failures (2\% probability), and (3) cases where evidence references became null due to timing issues. For incident investigation, such decisions benefit from the high $ET_{partial}$ score (0.98), indicating that most of the reasoning chain remains documented even when complete traceability is not achieved.

\textbf{Temporal Continuity:} TRACE achieves 99.0\% temporal continuity. The 1.0\% gap (5 of 500 records) corresponds to decision cycles where audit record generation encountered write timeout issues during high-load scenarios. Post-hoc methods like LIME and SHAP achieve 0\% TC because they explain individual predictions without maintaining cross-decision state.

\textbf{Decision Reconstructability:} TRACE achieves 98.1\% reconstructability (23 of 500 records not fully reconstructable). The shortfall occurs when rationale gets truncated due to buffer overflow (2\%) or counterfactual generation times out (2.5\%). In these cases, the action and alternatives are still logged, but the complete justification chain is incomplete. Baseline methods do not emit decision alternatives or rationale by design.

\textbf{Per-Scenario Analysis:} Table~\ref{tab:scenario} shows the breakdown for TRACE. Evidence traceability varied from 96.4\% (clear path with sparse causal chains) to 100\% (forklift with straightforward sensor fusion). TC and DR showed scenario-dependent variations reflecting computational load and decision complexity.

\begin{table}[t]
\centering
\caption{Per-Scenario Auditability (TRACE, 100 cycles each)}
\label{tab:scenario}
\renewcommand{\arraystretch}{1.15}
\begin{tabular}{@{}lcccc@{}}
\toprule
\textbf{Scenario} & \textbf{ET} & \textbf{ET}$_{partial}$ & \textbf{TC} & \textbf{DR} \\
\midrule
clear\_path & 96.4\% & 0.98 & 100\% & 96.6\% \\
ped\_crossing & 99.0\% & 0.98 & 100\% & 98.2\% \\
ped\_close & 99.5\% & 0.99 & 100\% & 99.0\% \\
multi\_ped & 98.3\% & 0.98 & 100\% & 98.0\% \\
forklift & 100\% & 1.00 & 99\% & 98.8\% \\
\bottomrule
\end{tabular}
\end{table}

\subsection{Example Decision Trace}

The pedestrian close scenario demonstrates TRACE's auditability. When a worker was detected at 1.5m, the system selected STOP. The logged rationale stated: ``Pedestrian detected at 1.5m with collision risk 0.70.'' The counterfactuals indicated: ``Would SLOW if distance $\geq$ 2.0m'' and ``Would CONTINUE if collision\_risk $<$ 0.2.'' The causal graph recorded: LiDAR scan $\rightarrow$ point cloud cluster $\rightarrow$ pedestrian classification $\rightarrow$ distance estimate $\rightarrow$ risk computation $\rightarrow$ action selection.

\section{Regulatory Alignment}

TRACE directly addresses auditability requirements emerging from AI regulation, providing concrete mechanisms for compliance verification.

\textbf{EU AI Act Compliance.} The EU AI Act (Regulation 2024/1689) establishes comprehensive requirements for high-risk AI systems, which include autonomous robots operating in workplaces. Article 12 requires that high-risk AI systems maintain logs enabling tracing of system functioning throughout their lifecycle. TRACE's complete temporal continuity and documented evidence chains satisfy this requirement by producing immutable audit records for every decision cycle. Article 13 specifies that systems must be designed to ensure sufficient operational transparency for users to interpret outputs appropriately. TRACE's high evidence traceability demonstrates that decisions can be traced to specific sensor inputs, enabling meaningful interpretation. Article 14 requires that operators be able to correctly interpret the system's output and decide whether to use, override, or reverse it. TRACE's decision reconstructability ensures every decision can be understood in terms of the action taken, alternatives considered, and rationale for selection.

\textbf{ISO 13482 Compliance.} ISO 13482 for personal care robot safety requires documented risk assessment procedures. Traditional neural network approaches cannot provide the decision-level documentation required for rigorous safety analysis. TRACE's audit trails provide the evidence basis for such assessments, enabling investigators to reconstruct decisions after incidents and identify systematic failure patterns that inform risk mitigation strategies.

\textbf{Certification Implications.} Safety certification bodies increasingly require evidence that autonomous systems can explain their decisions. TRACE provides the artifact structure necessary for third-party audits, where independent reviewers can verify that logged decisions align with safety specifications without requiring access to proprietary model weights or training data.

\section{Discussion}

Our results demonstrate that architectural auditability substantially outperforms post-hoc methods. TRACE achieves near-complete auditability while post-hoc baselines provide limited evidence traceability and lack decision-level artifacts.

The gap is architectural rather than algorithmic. Post-hoc methods explain model outputs while TRACE documents decisions. LIME, SHAP, and attention were designed to interpret neural network predictions, not to maintain audit trails for accountability. This distinction has practical implications: when an incident occurs, investigators need to reconstruct the sequence of decisions and understand why the robot chose specific actions over alternatives. Post-hoc methods cannot provide this capability because they operate on individual predictions rather than decision sequences.

\textbf{Design Trade-offs.} TRACE introduces explicit overhead for audit trail maintenance. The 0.1ms per decision cycle represents approximately 1\% of the available computation budget at 10Hz control rates. This overhead scales linearly with the number of entities and candidate actions, which may become significant in environments with many dynamic objects. However, the overhead remains bounded by the fixed decision cycle rate.

\textbf{Integration with Existing Systems.} TRACE is designed as a wrapper architecture that can integrate with existing perception and planning modules. The key requirement is that underlying modules expose their intermediate representations (detected entities, belief updates, candidate actions) rather than only final outputs. Most modern robotic software stacks already maintain such representations for debugging purposes.

\textbf{Generalization Beyond Warehouses.} While our evaluation focused on warehouse navigation, the TRACE architecture applies to any domain where autonomous systems make safety-critical decisions. Autonomous vehicles, surgical robots, and agricultural equipment all face similar auditability requirements. The specific sensor modalities and action spaces differ, but the four-layer structure and auditability metrics transfer directly.

\textbf{Limitations.} Several limitations warrant discussion. First, evaluation used simulated sensors with idealized noise models; real hardware introduces sensor synchronization challenges, communication delays, and failure modes not captured in simulation. Second, the reported timing (0.1ms) was measured in a controlled simulation environment; actual overhead on embedded robotic hardware may differ significantly. Third, metrics measure reconstruction capability from logged artifacts but do not validate whether investigators find reconstructions useful in practice. Fourth, the 2\% of decisions with incomplete evidence traceability may include safety-critical edge cases. Fifth, storage projections are theoretical estimates that require validation at fleet scale. Finally, the warehouse navigation domain, while representative, does not capture the full complexity of other robotic applications such as surgical or agricultural robots.

\textbf{Future Work.} We plan ROS 2 implementation for physical validation on mobile robot platforms, user studies with incident investigators to validate practical utility, formal verification integration to prove safety properties from audit trails, and advanced compression algorithms optimized for TRACE artifact structure to enable cost-effective fleet-scale deployment.

\section{Conclusion}

This paper presented TRACE, a decision framework designed to make autonomous robot behavior auditable through architectural mechanisms rather than post-hoc analysis. Experimental evaluation using a complete simulation with realistic failure modes (500 decision cycles across 5 scenarios) demonstrated 98.6\% evidence traceability, 99.0\% temporal continuity, and 98.1\% decision reconstructability.

Post-hoc XAI methods provide feature attributions but lack the artifact structure needed for decision-level reconstruction, representing a fundamental architectural distinction rather than a performance gap. TRACE addresses EU AI Act requirements for high-risk AI system logging while achieving auditability with approximately 0.12ms overhead per decision cycle.

The evaluation was conducted entirely in simulation with modeled sensor noise and failure modes. The simulation code is available for reproducibility. Validation on physical hardware with real sensors remains necessary before deployment conclusions can be drawn.

\section*{Acknowledgment}

The author thanks colleagues in the IEEE community for discussions on trustworthy AI systems and autonomous robotics safety standards.

\end{document}